\documentclass[10pt,twocolumn,letterpaper]{article}

\usepackage{cvpr}              

\usepackage{graphicx}
\usepackage{amsmath}
\usepackage{amssymb}
\usepackage{booktabs}
\usepackage{mathtools}

\usepackage{times}
\usepackage{epsfig}
\usepackage{amsmath}
\usepackage{bbm}
\DeclareMathAlphabet\mathbfcal{OMS}{cmsy}{b}{n}

\usepackage{float}

\usepackage{lipsum}
\usepackage{stfloats}
\usepackage{multicol}
\usepackage{multirow}
\usepackage{bm}
\usepackage{etoolbox}
\usepackage{icomma} 
\usepackage{array}
\usepackage{tabulary}
\usepackage[table]{xcolor} 
\usepackage{paralist}
\usepackage{booktabs}
\usepackage{adjustbox}

\usepackage{caption}
\usepackage{subcaption}

\usepackage{arydshln}

\definecolor{gray}{rgb}{0.3,0.3,0.3}
\definecolor{blue}{rgb}{0,0.5,1}
\definecolor{mask_red}{rgb}{1,0,0.8}
\definecolor{green}{rgb}{0.2,1,0.2}
\definecolor{rblue}{rgb}{0,0,1}
\definecolor{lightblue}{HTML}{6495ed}
\definecolor{lightred}{HTML}{F19C99}

\definecolor{graytablerow}{gray}{0.6}

\definecolor{revised_color}{HTML}{0066CC}
\definecolor{revised_color_PKY}{HTML}{FF2E82}
\definecolor{revised_color_SH}{HTML}{007FFF}

\usepackage[pagebackref,breaklinks,colorlinks]{hyperref}
\hypersetup{colorlinks, citecolor=blue}

\usepackage[capitalize]{cleveref}
\crefname{section}{Sec.}{Secs.}
\Crefname{section}{Section}{Sections}
\Crefname{table}{Table}{Tables}
\crefname{table}{Tab.}{Tabs.}

\usepackage[accsupp]{axessibility}

\def\cvprPaperID{3} 
\def\confName{OmniCV}
\def\confYear{2023}

\makeatletter
\renewcommand*{\@fnsymbol}[1]{\ensuremath{\ifcase#1\or *\or \dagger\or \ddagger\or
    \mathsection\or \mathparagraph\or \|\or **\or \dagger\dagger
    \or \ddagger\ddagger \else\@ctrerr\fi}}
\makeatother

\begin{document}

\title{FishDreamer: Towards Fisheye Semantic Completion via Unified Image Outpainting and Segmentation
}

\author{Hao Shi$^{1,5,}$\thanks{The first two authors contribute equally to this work.}\, ,
~~Yu Li$^{2,*}$,
~~Kailun Yang$^{3,}$\thanks{Corresponding author (e-mail: {\tt kailun.yang@hnu.edu.cn, wangkaiwei@zju.edu.cn}).}\, ,
~~Jiaming Zhang$^{2,4}$,
~~Kunyu Peng$^{2}$,
~~Alina Roitberg$^{2}$,\\
Yaozu Ye$^{1}$,
~~Huajian Ni$^{5}$,
~~Kaiwei Wang$^{1,\dagger}$,
~~Rainer Stiefelhagen$^{2}$\\
\normalsize
$^{1}$Zhejiang University
~~$^{2}$Karlsruhe Institute of Technology
~~$^{3}$Hunan University
~~$^{4}$University of Oxford
~~$^{5}$Supremind
}
\maketitle

\begin{abstract}
This paper raises the new task of \textbf{Fisheye Semantic Completion} (FSC), where dense texture, structure, and semantics of a fisheye image are inferred even \textit{beyond the sensor field-of-view} (FoV). Fisheye cameras have larger FoV than ordinary pinhole cameras, yet its unique special imaging model naturally leads to a blind area at the edge of the image plane. This is suboptimal for safety-critical applications since important perception tasks, such as semantic segmentation, become very challenging within the blind zone. Previous works  considered the out-FoV outpainting  and in-FoV segmentation separately. However, we observe that these two tasks are actually closely coupled. To jointly estimate the tightly intertwined complete fisheye image and scene semantics, we introduce the new \textbf{FishDreamer} which relies on successful ViTs enhanced with a novel Polar-aware Cross Attention module (PCA) to leverage dense context and guide semantically-consistent content generation while considering different polar distributions. In addition to the contribution of the novel task and architecture, we also derive Cityscapes-BF and KITTI360-BF datasets to facilitate training and evaluation of this new track. Our experiments demonstrate that the proposed FishDreamer outperforms methods solving each task in isolation and surpasses alternative approaches on the Fisheye Semantic Completion. Code and datasets are publicly available at \href{https://github.com/MasterHow/FishDreamer}{FishDreamer}.

\end{abstract}

\section{Introduction}
\label{sec:intro}
\begin{figure}[!t]
   \centering
   \includegraphics[width=0.95\linewidth]{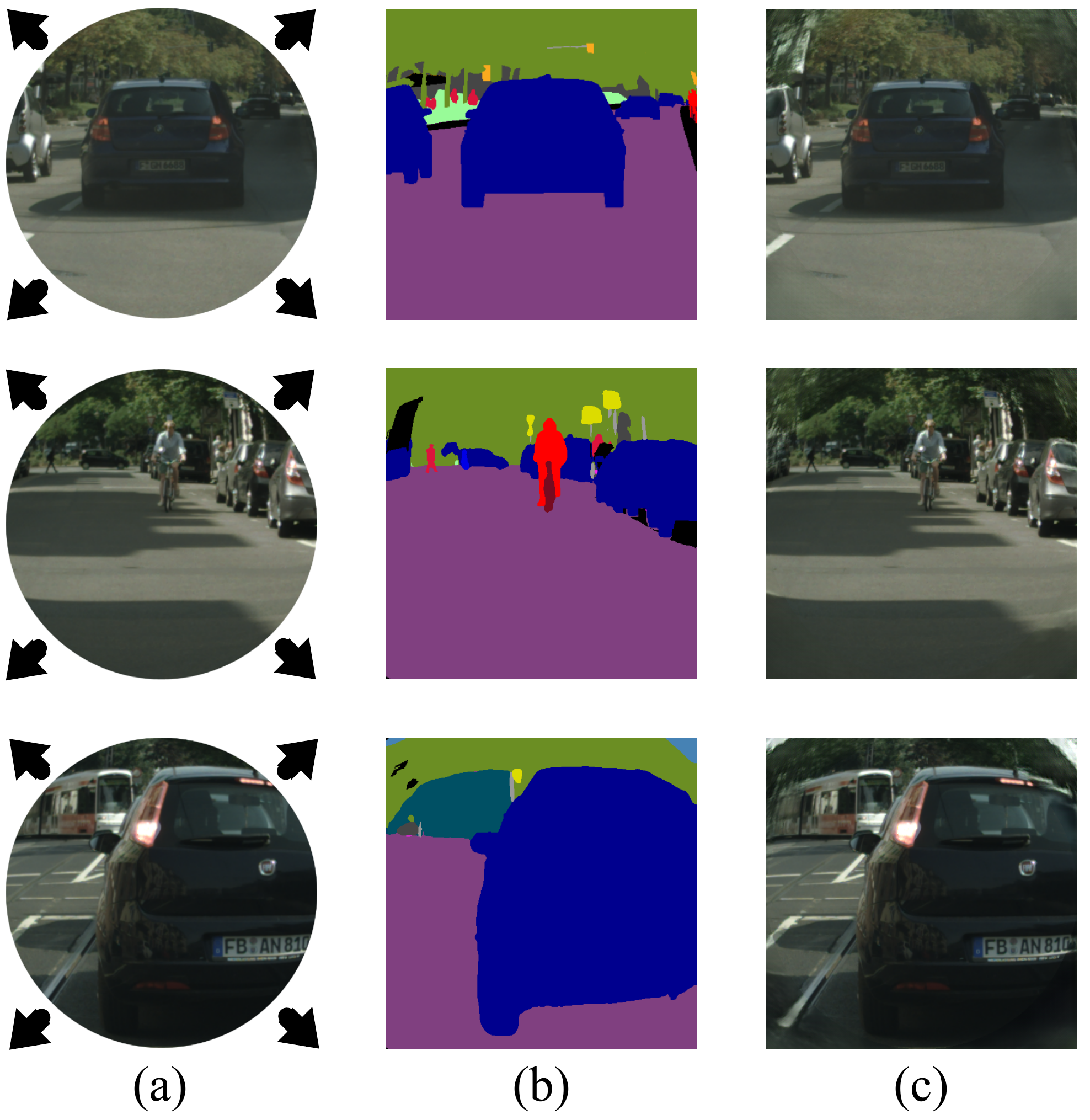}
   \vspace{-1em}
   \caption{\emph{Illustrations of our proposed \textbf{Fisheye Semantic Completion}.} (a) Input single-view fisheye image. (b)(c) Fisheye semantic completion result: our model jointly predicts texture, structure and object categories for each pixel beyond the fisheye vision.}
   \label{fig:teaser}
  \vspace{-1.0em}
\end{figure}

Benefiting from a larger field-of-view (FoV), fisheye cameras have been widely used in autonomous driving and mobile robots~\cite{yogamani2019woodscape,qian2022survey_fisheye,kumar2023surround_survey,gao2022review_panoramic_imaging,sekkat2020omniscape}. However, due to the special optical design of the fisheye camera, there are invalid black areas at the edge of the image plane. Interestingly, humans have a natural ability to infer complete semantic information from partial visual observations~\cite{pessoa2003filling,lin2022neural} (\eg, a partially occluded car) to navigate and interact in the real world. Similarly, for an ego-view agent, the ability to estimate the full field-of-view of a given scene is beneficial for mid-level tasks such as obstacle avoidance~\cite{kim2015rear}, while perceiving semantic concepts is a prerequisite for complex cognitive tasks such as high-level scene understanding, planning the next step or answering questions about the space~\cite{guerrero2020s, cartillier2021semantic}.

With this motivation, our goal is to build a model that can simultaneously complete the missing image areas and generate predictions for semantic object categories from a single-shot fisheye image in an end-to-end manner.
We refer to this novel task as ``Fisheye Semantic Completion'', with an overview of the proposed problem given in Fig. \ref{fig:teaser}.
Our key idea is grounded by the  observation  that the distribution of pixel values of an entity within an image is tightly coupled to its semantic label.
Therefore, the two problems of outpainting the content outside the fisheye camera's FoV and the semantics of objects are strongly conjugated, which we believe is  vital for achieving good performance in both tasks.
In other words, if we know the semantic categories of an incomplete object, we can predict its pixel pattern on the image plane even without direct observation (\eg, seeing a tree trunk appearing in the FoV and then inferring the the presence of tree canopy outside). 
Conversely, having a complete observation of an object can help us recognize its semantic class more accurately.

To achieve this goal, we must overcome several key challenges: First, how to effectively take advantage of the strong coupling characteristics of these two sub-tasks (\ie fisheye outpainting and segmentation) to realize a win-win situation? Second, since existing fisheye semantic segmentation datasets cannot provide images and semantic ground truth outside the FoV, how can we obtain fisheye beyond-FoV training data with complete annotations?

To address the first challenge,  we propose \textbf{FishDreamer}, which benefits from the successful Visual Transformer (ViT) structure as the backbone, and integrates a novel Polar-aware Cross Attention module (PCA) to enhance the flow of visual cues between two sub-tasks.
Specifically, PCA takes into account the unique polar distribution and distortion patterns of fisheye cameras, considers the heterogeneity of different polar coordinate locations when querying relevant visual features, and leverages the rich semantic context from the semantic head to guide the outpainting head in hallucinating semantically continuous and plausible content outside the fisheye FoV.
As for the data challenge, we leverage the popular Cityscapes~\cite{cordts2016cityscapes} and KITTI360~\cite{liao2022kitti} semantic segmentation datasets via pinhole-to-fisheye projection and derive the new CityScapes-BF and KITTI360-BF as beyond-FoV benchmark variants, therefore enabling training and evaluation of fisheye semantic completion.

Extensive experiments demonstrate that the proposed FishDreamer, which jointly learns semantics and content outside the fisheye FoV, outperforms approaches that address the two sub-tasks in isolation. 
The proposed PCA module, which focuses on the natural polarity distribution of the fisheye image and extracts visual cues from semantic priors, significantly improves performance.
On the derived CityScapes-BF benchmark, FishDreamer achieves state-of-the-art performance with a mIoU of $54.54\%$ and a peak-SNR of $25.05dB$, a $0.42dB$ performance gain from the best published result ($24.63dB$).
FishDreamer also surpasses alternative approaches on KITTI360-BF as it hallucinates more realistic content in the blind area of the fisheye and gives clearer and sharper segmentation results.

In summary, we deliver the following contributions:
\begin{compactitem}
    \item We raise the new Fisheye Semantic Completion task, which extends beyond fisheye vision and enables  outpainting and semantic segmentation of the full scene.
    
    \item We establish the CityScapes-BF and KITTI360-BF benchmarks and validate existing models on this  new fisheye semantic completion track.

    \item We propose FishDreamer, which utilizes a novel Polar-aware Cross Attention (PCA) module to effectively guide fisheye outpainting using semantic context.

    \item  Extensive experiments demonstrate that the proposed FishDreamer outperforms alternative approaches that address the sub-tasks separately.
\end{compactitem}

\section{Related Work}
\label{sec:related_work}
\noindent\textbf{Image outpainting.} 
Image outpainting aims to generate the surrounding regions of the given visual content.
Early parameter-free methods~\cite{wang2014biggerpicture,zhang2013framebreak,shan2014photo} are data-driven and are based on very large image databases or require input reference frames, which retrieve relevant image features to warp and fill in  regions-of-interest.
Sabini~\etal~\cite{sabini2018painting} first present the learning-based image outpainting via Generative Adversarial Network (GAN)~\cite{goodfellow2020generative} to enable outpainting in the horizontal direction.
Subsequently, Wang~\etal~\cite{wang2019wide} proposes generating semantically coherent structures and textures using a context prediction network and a carefully designed loss function.
The framework of Teterwak~\etal\cite{teterwak2019boundless}, leverages semantic information extracted from a pretrained deep network to modulate the discriminator's behavior for image extension. 
Yao~\etal~\cite{yao2022outpainting} implemented a sequence-to-sequence outpainting approach that relies on a transformer-based backbone, where the outpainting proportion and the network structure are bound.
FlowLens~\cite{shi2022flowlens} introduces a temporal clip propagation mechanism to expand the FoV of the pinhole camera outwards and the spherical camera inwards, respectively.
RecRecNet~\cite{liao2023recrecnet} rectangles rectified wide-angle via curriculum learning with increasing degree of freedom.
The work most closely related to ours is presumably FisheyeEX~\cite{liao2022fisheyeex}, which leverages an outpainting method specifically for elimination of fisheye blind areas.

Different from prior works, we focus on \textit{fisheye semantic completion}, that seeks to generate \textit{semantically coherent}  visual content beyond the fisheye FoV by jointly learning the scene semantics and pixel patterns and considering the fisheye polar distributions.
To the best of our knowledge, this is the first work that concurrently addresses the challenges of fisheye semantic segmentation and scene completion.

\noindent\textbf{Beyond-FoV semantic segmentation.}
Early fisheye semantic segmentation methods~\cite{deng2017cnn,saez2018cnn,sekkat2022comparative} generate synthetic fisheye datasets based on existing pinhole semantic segmentation datasets.
These techniques employ focal length augmentation to enhance their adaptability to real-world scenarios.
In~\cite{blott2018semantic,ye2020universal}, the degree of freedom in generating synthetic fisheye images is enlarged, transforming rectilinear images to fisheye images in a more comprehensive way and enhancing the generalization on real fisheye images with various perspectives.
In~\cite{ahmad2022fisheyehdk,deng2019restricted,hu2022distortion_convolution,playout2021adaptable_deformable_convolutions}, deformable components in CNNs are investigated to better adapt to wide-angle images.
In~\cite{yang2019can,yang2019pass,yang2020dspass,xu2019semantic_synthetic,orhan2022semantic_outdoor}, the FoV is further expanded to 360{\textdegree} with panoramic or annular images.
In~\cite{yang2021capturing,yang2021context,zhang2022bending,zhang2022behind}, wide-FoV-driven visual attention and distortion-aware transformer models are designed to learn long-range dependencies in panoramic images.
In~\cite{kim2022pasts,yang2020omnisupervised,zheng2023complementary}, knowledge distillation is studied on panoramic images.
In~\cite{jang2022dada,ma2021densepass,kim2022pasts,shi2022unsupervised,zhang2021transfer}, large-FoV semantic segmentation is revisited from a domain adaptation perspective by adapting from label-rich pinhole images to label-scare images such as fisheye images, panoramic images, and images reflected by convex mirrors.
In~\cite{jaus2021panoramic,mei2022waymo,thioune2022fpdm}, semantic segmentation is extended to panoptic segmentation on wide-FoV images with instance predictions.
In~\cite{cheke2022fisheyepixpro,jaus2021panoramic,jaus2023panoramic}, pixel-level contrastive learning is studied for wide-angle segmentation.
In~\cite{arsenali2019rotinvmtl,eising2021near_field_perception,kumar2021syndistnet,kumar2021omnidet}, multi-task learning has been implemented on fisheye images such as object detection, depth estimation, and semantic segmentation.
In contrast to these works, our work tackles fisheye semantic completion, which provides dense semantic information not only for the original wide-angle fisheye images but also beyond the field of view, giving rich semantic understanding of the expanded scene. 

\section{Methodology}
\label{sec:methodology}
\subsection{Overview}
\begin{figure}[!t]
   \centering
   \includegraphics[width=0.925\linewidth]{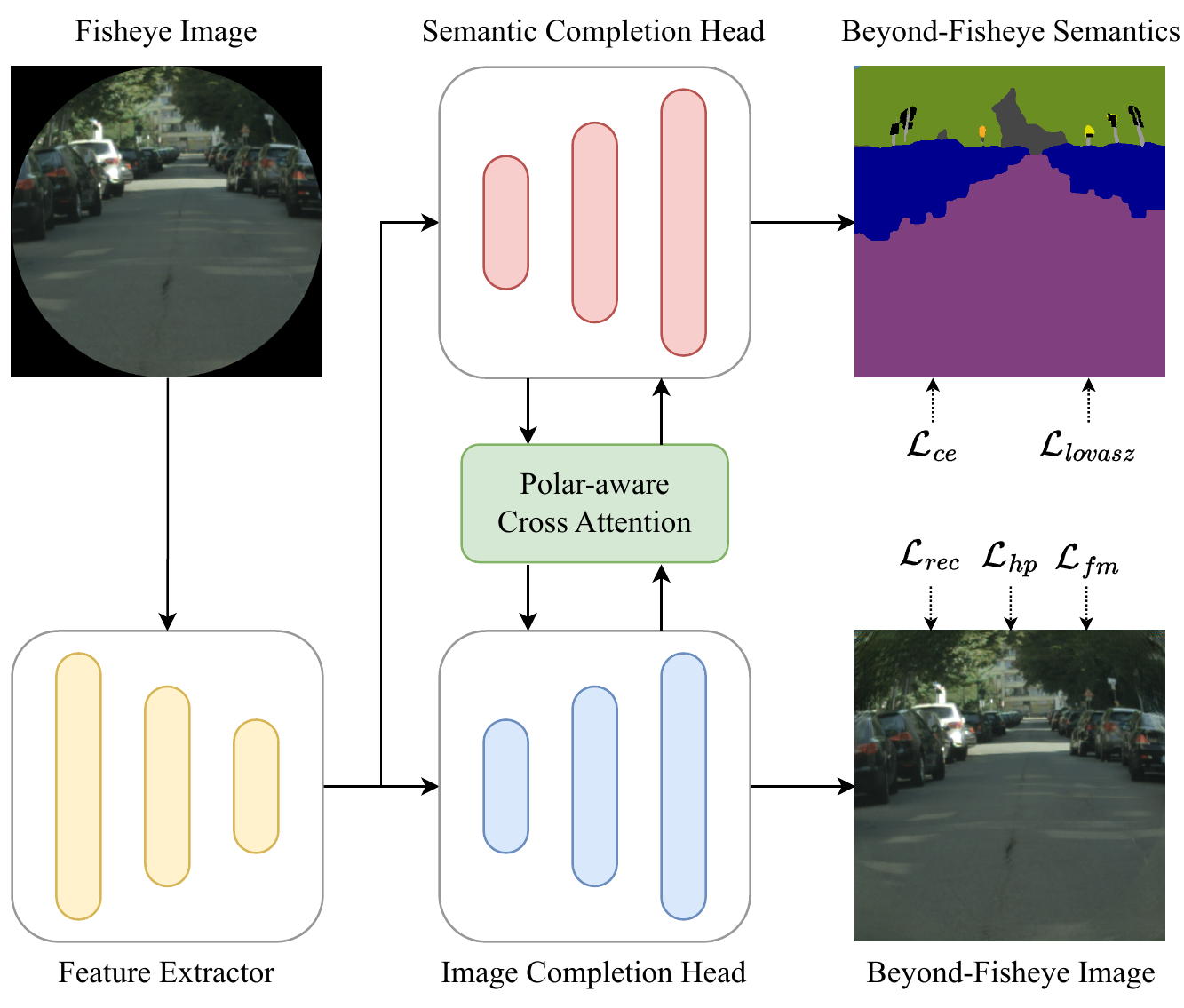}
  \vspace{-1em}
   \caption{\emph{FishDreamer: Fisheye semantic completion transformer.} Taking a single fisheye image as input, the model predicts pixel patterns and object labels beyond fisheye vision. Swin~\cite{liu2021swin} is used as the backbone. The two sub-task heads interact each other via the proposed Polar-aware Cross Attention module bidirectionally.}
   \label{fig:overview}
  \vspace{-1.5em}
\end{figure}

In this section, we introduce FishDreamer -- a novel approach capable of simultaneously achieving reliable extrapolation of fisheye images and a full-FoV semantic segmentation for both visible and previously unseen areas within fisheye images. 
As illustrated in Fig.~\ref{fig:overview},  FishDreamer comprises four modules: the feature extraction backbone  based on the Swin Transformer~\cite{liu2021swin} (Sec.~\ref{sec:feature_extractor}), the outpainting module (Sec.~\ref{sec:outpainting}), the semantic decoder based on UPerNet~\cite{xiao2018unified} (Sec.~\ref{sec:segmentation}) and the Polar-aware Cross Attention (PCA) mechanism (Sec.~\ref{sec:polar_aware_cross_attention}).
Next, we will provide a comprehensive description of each module and their respective roles within the FishDreamer framework. 

\subsection{Feature Extractor}
\label{sec:feature_extractor}

We begin by describing our feature extraction backbone, the Swin Transformer~\cite{liu2021swin}, which facilitates informative hierarchical feature learning and has been proven very effective, \eg, in image and video classification~\cite{liu2021swin, liu2022video}, activity recognition~\cite{peng2022transdarc} and vanilla semantic segmentation~\cite{lin2022ds}. 
Similar to other transformer-based models Swin leverages self-attention~\cite{vaswani2017attention}, but also employs a non-overlapping shifted window partitioning mechanism which enhances efficiency by focusing on the generated windows while preserving cross-window communication capabilities.

We utilize a four-stage Swin Transformer as our feature extraction backbone. 
In the first stage, multiple non-overlapping image patches are generated. As we address outpainting and semantic segmentation using 2D image data as input, a 2D shifted window pipeline is employed, operating within the 2D spatial domain.
Assuming the 2D spatial dimensions of the input image are $[H, W]$ and the shifted window size is chosen as $[N, N]$, a total of $\frac{H}{N}\times\frac{W}{N}$ patches are extracted using the aforementioned window partitioning technique. Next, these patches are projected from $\textsc{R}^{3}$ to $\textsc{R}^{C}$ using a linear projection layer. Next, we will describe further details of the Swin Transformer blocks .

The Swin Transformer block leverages its own  Shifted-Window based Multi-head Self-Attention (SW-MSA) mechanism, as opposed to the standard Multi-head Self-Attention (MSA) found in ViT~\cite{dosovitskiy2020image}. This approach mitigates the limitations of the vanilla ViT structure, specifically concerning the lack of cross-window connections and restricted model capacity. The workflow of the Swin Transformer block can be described as:
\begin{equation}
\begin{split}
    \hat{z}_{l-1} = SW-MSA(LN(z_{l-1})),\\
    z_{l} = MLP(LN(\hat{z}_{l-1}) + \hat{z}_{l-1},
\end{split}
\end{equation}
where SW-MSA stands for the Shifted-Window based Multi-head Self-Attention, LN indicates layer normalization, MLP is multi layer perception with GELU nonlinearity, $l$ marks the layer number, $\hat{z}_{l-1}$ is the attention output, $z_{l-1}$ and $z_{l}$ are the outputs of $\{l-1\}{-}th$ and $\{l\}{-}th$ modules, respectively.
After each module there is a residual connection.
All Swin blocks are equipped with the shifted window partitioning approach described above.
Swin processes the image in a hierarchical manner, since it splits the input into non-overlapping patches and subsequently merges them at different resolutions.

\subsection{Outpainting}
\label{sec:outpainting}
Outpainting approaches typically employ several layers of deconvolution as the task head, and FishDreamer is no exception. 
Given that fisheye semantic completion seeks to accomplish both image completion and semantic completion simultaneously, we opt not to incorporate a more complex outpainting decoder in the design.
Specifically, FishDreamer uses three layers of deconvolution, along with the PCA mechanism described in Sec. \ref{sec:polar_aware_cross_attention}, to upsample and produce the feature map obtained from the feature extractor. This process generates the extrapolated output, which can be calculated as follows:
\begin{equation}
\centering
\begin{split}
    \hat{z_s} = ConvT_{i}(z_s),\\
    z = ConvT_{i+1}(\hat{z_s} +PCA(\hat{z_s},z_c)),\\
    z := ConvT_{i+2}(z),
\end{split}
\end{equation}
where ConvT is the 2D transposed convolution operator with kernel size$=3\times3$ while the stride and padding are set as 2 and 1. $i$ indicates the stage of the ConvT and is chosen as 0 in this work. PCA is the proposed Polar-aware Cross Attention module which will be detailed in Sec.~\ref{sec:polar_aware_cross_attention}, $z_{s}$ and $z_c$ denote the feature extracted from the Swin Transformer-based backbone and the features from semantic completion head, $\hat{z}$ denotes the feature map of the semantic segmentation branch after the first ConvT layer, $z$ denotes the final merged feature map, respectively.
With  additional priors from the semantic completion head we can acquire a semantically coherent outpainting result.

\subsection{Segmentation}
\label{sec:segmentation}
\begin{figure}[!t]
   \centering
   \includegraphics[width=1.0\linewidth]{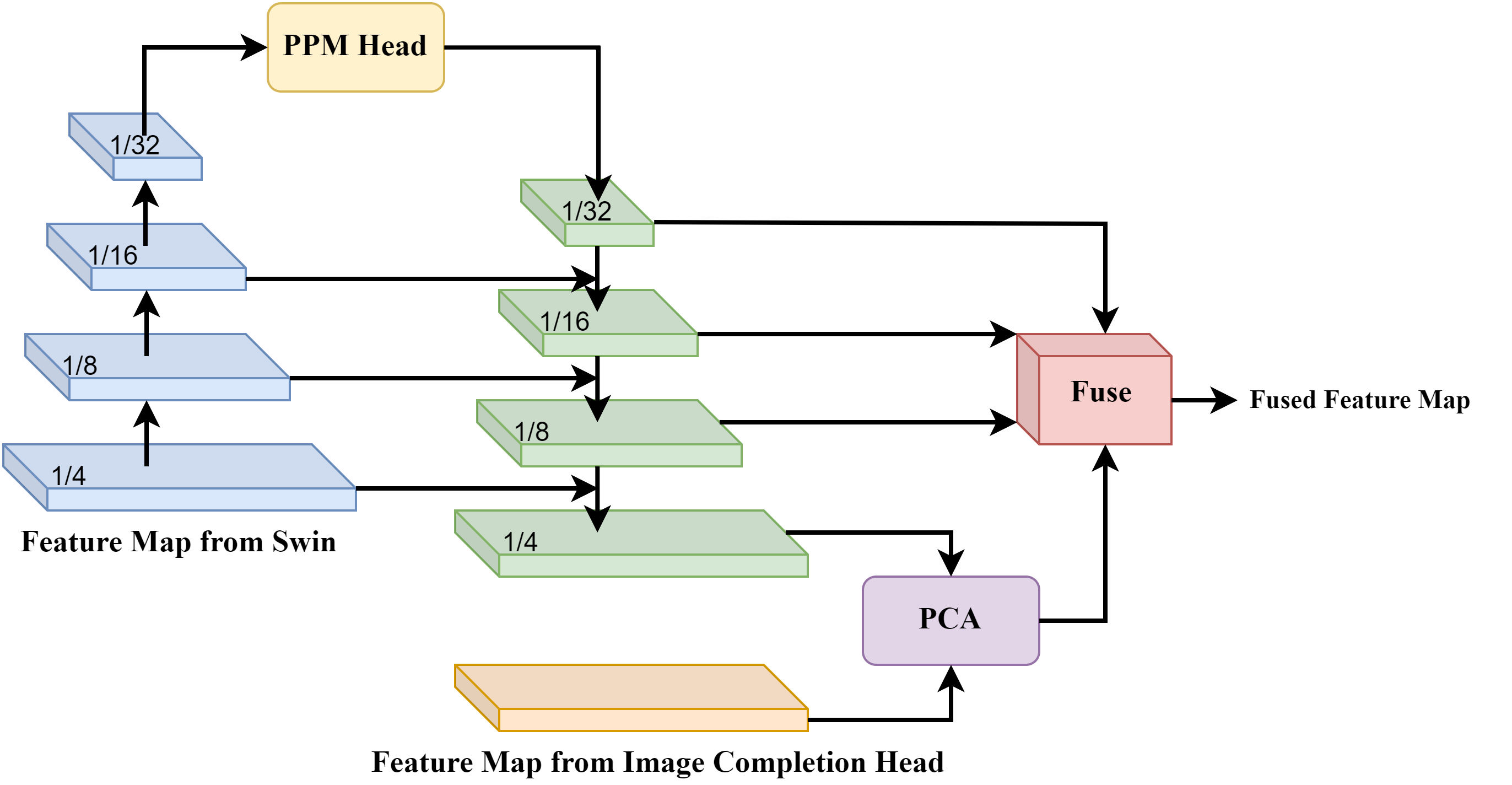}
   \caption{\emph{UPerNet with PCA.} Hierarchical UPerNet with extra information from the image completion head.}
   \label{fig:upernet}
  \vspace{-1.0em}
\end{figure}
Considering the hierarchical structure of the features obtained using the Swin Transformer backbone, more advanced techniques for dense prediction, such as Feature Pyramid Networks (FPN)~\cite{lin2017feature} and U-Net~\cite{ronneberger2015u}, can be employed for linking the hierarchical information. 
In our work, we incorporate UPerNet~\cite{xiao2018unified} with our proposed PCA.

As depicted in Fig.~\ref{fig:upernet}, UPerNet features a top-down architecture with lateral connections that facilitate the fusion of high-level semantic information with lower-level details. The model utilizes a Pyramid Pooling Module (PPM)~\cite{zhao2017pyramid} to achieve a larger receptive field and generate effective global prior representations.
UPerNet has the capability to learn visual attributes for semantic segmentation and image completion at multiple levels.
$\{F_{1}, F_{2}, F_{3}, F_{4}\}$ denotes the set of the resulting feature maps of each stage of the backbone.
The corresponding downsampling rate for the $\{F_{1}, F_{2}, F_{3}, F_{4}\}$ are $\{4,8,16,32\}$, respectively.
In the decoder stage,  the PPM is only used at the top of $F_{4}$, where the resulting feature map can be denoted as $\hat{F}_{4}$.
Then, $\hat{F}_{4}$ is upsampled and sum with $F_{3}$ to progressively fuse multi-level features. The upsampling and downsampling rates are kept the same and finally four feature maps are obtained through the decoder stage.
The corresponding feature maps for the four stages of the decoder are be referred to as $\{\hat{F}_{1}, \hat{F}_{2}, \hat{F}_{3}, \hat{F}_{4}\}$. $\hat{F}_{4}$ (which is alternatively denoted as $z_s$) is then merged with the feature $z_c$ from the image completion branch by using the PCA module explained in the next subsection.
The final feature map can be obtained as follows: 
\begin{equation}
    \centering
    z_{out} = Fuse(\hat{F}_{1}, \hat{F}_{2}, \hat{F}_{3}, z_{pca}),
\end{equation}
where the $z_{pca}$ denotes the output feature map of the PCA mechanism. We fuse the feature maps via concatenation.

\subsection{Polar-aware Cross Attention}
\label{sec:polar_aware_cross_attention}
\begin{figure*}[!t]
   \centering
   \includegraphics[width=1.0\linewidth]{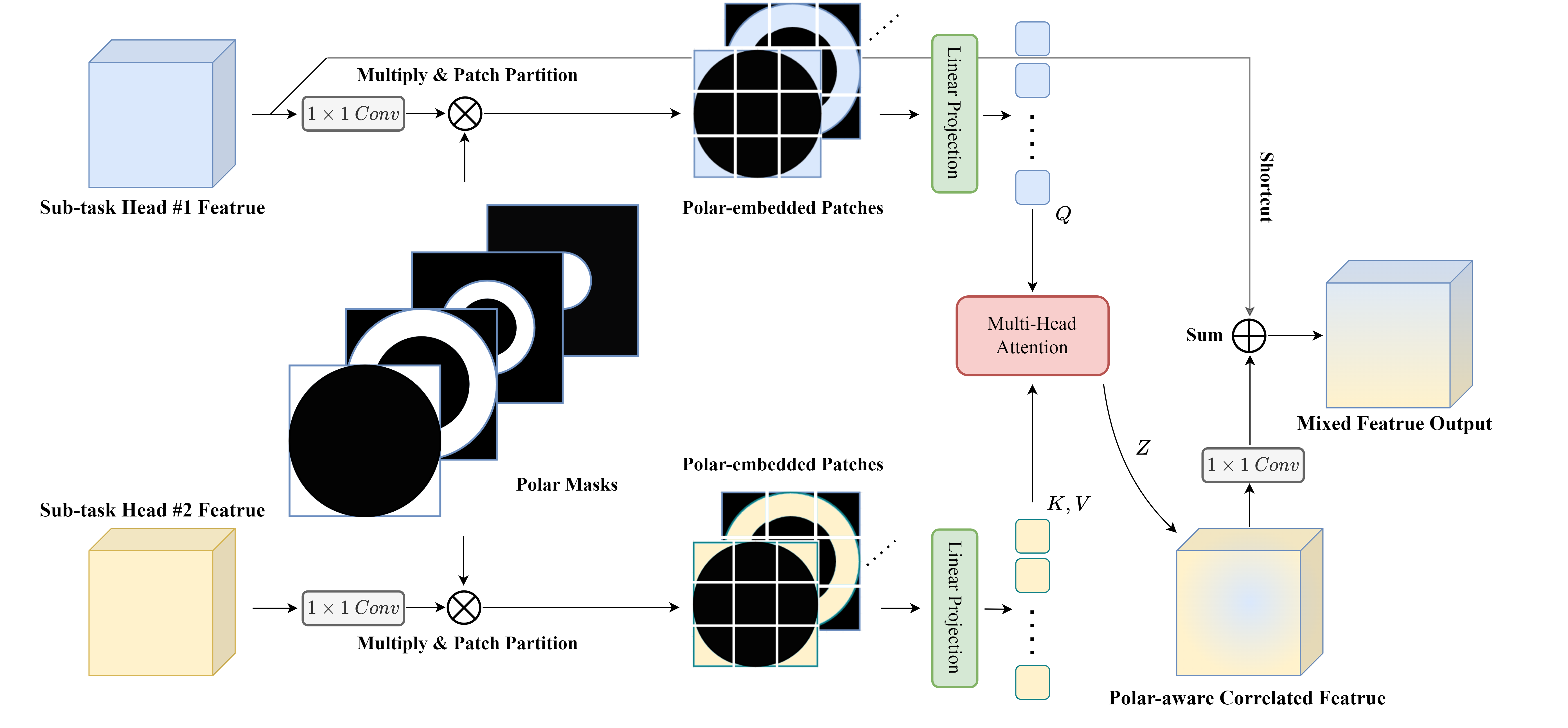}
   \caption{\emph{Polar-aware Cross Attention module (PCA)}. Given the feature maps from two sub-task heads, PCA explicitly limits the polarity distribution in each patch by introducing $N$ ring-shape masks, and exploit the multi-head attention to query polar-aware correlated feature from coupled sub-task feature map. Note this behavior can be either unidirectional or bidirectional.}
   \label{fig:pca}
\end{figure*}

As image outpainting and semantic segmentation tasks are tightly intertwined, we design the novel Polar-aware Cross Attention  (PCA) module to encourage the information flow between the two heads.

Given the two feature maps from the semantic segmentation head ($z_{s}$) and the image completion head ($z_{c}$), PCA initially constrains the polar distribution within each patch using a polar mask obtained from a newly designed polar-mask generator. This generator creates polar masks with varying quantities and radius.
Let the total number of the generated masks be $N_{mask}$ and the set of polar masks be $M =\{M_{i}~|~ i ~\in~ [0, N_{mask}]\}$. A linear projection layer is employed for the masked feature map of each sub-task. The projected and masked patch partitions for both the feature maps from the semantic segmentation- ($z_{s}^{}$) and the image completion heads ($z_{c}^{}$) can be computed as follows:
\begin{equation}
  z_{s}^{*}, z_{c}^{*} = L_{s}(M\odot z_{s}),  L_{c}(M\odot z_{c}),
\end{equation}
where the $L_{s}$ and $L_{c}$ indicate the linear projection layers of the semantic segmentation branch and the image completion branch, respectively.
A multi-head cross-attention mechanism is utilized to integrate the focuses from the image completion branch into the semantic segmentation branch. To this intent, linear projection layers $P_{Q}$, $P_{K}$, and $P_{V}$ are employed to compute the necessary query, key, and value components for the MSA. The resulting merged feature map is then passed through an additional bottleneck layer to obtain the combined feature map. This final merged feature map is subsequently added to $z_{s}^{*}$ to produce the ultimate output.
This workflow can be formalized as:
\begin{equation}
     \centering
     z_{pca} = z_{s}^{*}+C_{BN}(SA(P_{Q}(z_{s}^{*}),P_{K}(z_{c}^{*}), P_{V}(z_{c}^{*})),
\end{equation}
where $z_{pca}$ denotes the final mixed feature output of the proposed PCA mechanism, the $C_{BN}$ denotes the bottleneck layer and SA denotes the self-attention mechanism which is calculated as $SA(Q, K, V){=}Softmax(QK^{T}/\sqrt{{\rho}})V$ and $\rho$ denotes the scale factor~\cite{dosovitskiy2020image}.

In comparison to the previous FishFormer work~\cite{yang2022fishformer}, which solely focuses on one task, \ie, fisheye distortion correction, our model (shown in Fig.~\ref{fig:pca}) simultaneously addresses two critical tasks for autonomous driving, namely semantic segmentation and image completion, which involves predicting the unseen regions of fisheye images by considering the entire scene. 
This blind zones information is crucial for autonomous vehicles in order to minimize potential risks, such as route planning and risk alerting within the blind zones of fisheye sensors.
Our PCA mechanism is employed  at the end of the model, merging feature maps from both the semantic segmentation and image completion heads.
In contrast, the Layer Attention Mechanism (LAM) proposed by FishFormer, is integrated between every two transformer blocks, which is less efficient in terms of the number of attention blocks used. Additionally, while FishFormer encodes different annular slices, we force each token to encode features of a specific polar distribution. Tokens lacking valid features are discarded to enhance computational efficiency.  

\begin{figure*}[h]
  \centering
  \includegraphics[width=1\linewidth]{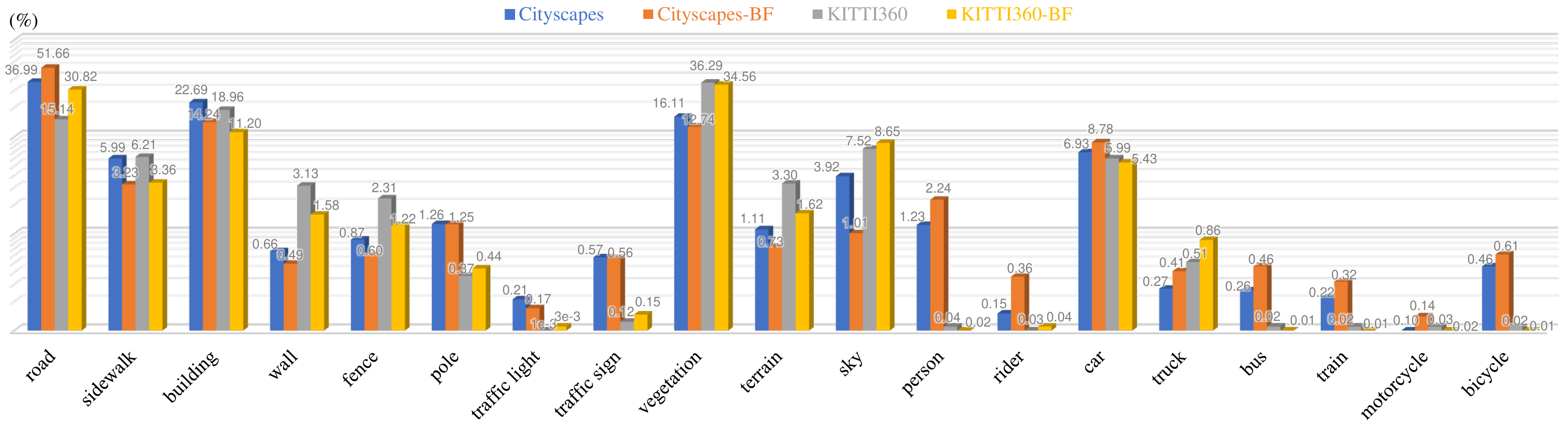}
  \vspace{-2em}
  \caption{\emph{Distributions of Cityscapes~\cite{cordts2016cityscapes}, Cityscapes-BF, KITTI360~\cite{liao2022kitti}, and KITTI360-BF in terms of class-wise pixel percentage across the datasets.} We use the logarithmic scaling of the vertical axis and insert the pixel frequency above the bar.}
  \label{fig:city_statis}
  \vspace{-1.0em}
\end{figure*}
\subsection{Training}
\noindent \textbf{Loss function.} 
Multiple well-established loss functions are employed to simultaneously ensure the accuracy of blind-area prediction and complete-FoV semantic segmentation.
Considering the full-scene image completion task of the fisheye image, we make use of a high receptive field perceptual loss, \ie, $ \mathcal{L}_{hp}$, an adversarial loss, \ie,  $\mathcal{L}_{adv}$, a reconstruction loss, \ie, $\mathcal{L}_{rec}$  and a feature matching loss, \ie, $\mathcal{L}_{fm}$.
Next, we discuss these loss functions in detail.

First, leverage the high receptive field perceptual loss~\cite{suvorov2022resolution} $\mathcal{L}_{hp}$, which calculates the difference between the feature maps of the predicted results and the target images.
This loss function does not require an exact reconstruction workflow, which is particularly suitable for our case when addressing the challenge of limited information in the blind area.
$\mathcal{L}_{hp}$ can be calculated as:
\begin{equation}
    \mathcal{L}_{hp}(z_p, z_t)=\mathcal{M}\left(D(\phi(z_p),\phi(z_t))^{2}\right),
\end{equation}
where $D(\cdot)$ denotes an element-wise distance function (mean-squared error) loss and $\mathcal{M}$ denotes the sequential two-stage mean operation.
The $\mathcal{L}_{hp}$ calculates the distance between the extracted features of the prediction ($z_p$) and the ground truth ($z_t$).
$\phi$ denotes dilated convolutions. 
$\mathcal{L}_{hp}$ does not require an exact reconstruction, which is a very good property in our case due to the lack of information of the blind area. 

Second, an adversarial loss is used to ensure the preservation of local details. A discriminator, $D_{\xi}(\cdot)$, is used to distinguish between "real" and "fake" patches. The visible parts of the built images are marked as real, while patches that intersect with the blind area are marked as fake. We then compute the non-saturating adversarial loss $\mathcal{L}_{adv}$ as:
\begin{equation}
\begin{aligned}
    \mathcal{L}_{D}=-\mathbb{E}_{x}[ {\left[\log D_{\xi}(x)\right]-\mathbb{E}_{x, m}\left[\log D_{\xi}(\hat{x}) \odot M\right] } \\
 -\mathbb{E}_{x, m}\left[\log \left(1-D_{\xi}(\hat{x})\right) \odot(1-M)\right],    
\end{aligned}
\end{equation}
\begin{equation}
     \mathcal{L}_{G}=-\mathbb{E}_{x, m}\left[\log D_{\xi}(\hat{x})\right],
\end{equation}
\begin{equation}
    L_{adv}=\operatorname{sg}_{\theta}\left(\mathcal{L}_{D}\right)+\operatorname{sg}_{\xi}\left(\mathcal{L}_{G}\right) \rightarrow \min _{\theta, \xi},
\end{equation}
where $x$ denotes a sample from dataset, $\hat{x}$ is the outpainting prediction, and $M$ denotes the circular masks to synthesized fisheye images. $\operatorname{sg}_{k}$ stops gradient \wrt \textit{k}, and $\mathcal{L}_{adv}$ is the joint adversarial loss which needs to be optimized. 

Third, $L_{rec}$ measures the L1 distance between the outpainted  image generated by FishDreamer and the ground truth, which can be calculated via the following equation:
\begin{equation}
    \mathcal{L}_{rec}=\Vert\hat{x}-x\Vert_1.
\end{equation}
Then, $L_{fm}$~\cite{wang2018high} is leveraged to denote a discriminated-based perceptual loss, which stabilizes training and improves the performance.

Finally, for the semantic completion we employ further two additional loss functions, \ie,  a cross entropy loss $ \mathcal{L}_{ce}$ and a Lov{\'a}sz-softmax loss $ \mathcal{L}_{lovasz}$~\cite{berman2018lovasz}.

The final FishDreamer loss $\mathcal{L}$ becomes:
\begin{equation}
\begin{aligned}
    \mathcal{L} &= \lambda_{adv} \cdot L_{adv} + \lambda_{hp} \cdot \mathcal{L}_{hp} + \lambda_{fm} \cdot \mathcal{L}_{fm} \\
    &+ \lambda_{rec} \cdot \mathcal{L}_{rec} + \lambda_{ce} \cdot \mathcal{L}_{ce} + \lambda_{lovasz} \cdot {L}_{lovasz},
\end{aligned}
\end{equation}
which is the weighted sum of the above losses. We empirically set $\lambda_{adv}{=}20, \lambda_{hp}{=}60, \lambda_{fm}{=}200, \lambda_{rec}{=}20, \lambda_{ce}{=}30$, and $\lambda_{lovasz}{=}10$ in all the experiments.

\begin{table}[!t]
\renewcommand{\thetable}{1}
    \begin{center}
        \caption{\emph{The training and validation sets splits of the derived Cityscapes-BF and KITTI360-BF.}}
        \label{tab:fov_expansion}
        \vspace{-1.0em}
\resizebox{1.0\columnwidth}{!}{
\setlength{\tabcolsep}{4mm}{ 
\begin{tabular}{l|ccc}
 \hline
    \textbf{Dataset} & \textbf{Train} & \textbf{Validation} & \textbf{Total} \\
 \hline\hline
    Cityscapes-BF & 2,975 & 500 & 3,475\\
    KITTI360-BF & 9,800 & 2,455 & 12,255 \\
 \hline
\end{tabular}
}
}

        \vspace{-2.25em}
    \end{center}
\end{table}
\section{Experiments}
\label{sec:experiments}

\subsection{Datasets}
To facilitate training and evaluation of fisheye semantic completion, we derive Cityscapes-BF and KITTI360-BF from the Cityscapes~\cite{cordts2016cityscapes} and KITTI-360~\cite{liao2022kitti} datasets.
To achieve this, we conduct a radial distortion of perspective images via the following equation~\cite{liao2020model}:
\begin{equation}
\begin{split}
    x_d = x_o(1 + k_1r^2_f + k_2r^4_f + k_3r^6_f + k_4r^8_f + ...) \\
    y_d = y_o(1 + k_1r^2_f + k_2r^4_f + k_3r^6_f + k_4r^8_f + ...),
\end{split}
\label{distortion-formula}
\end{equation}
where  $\boldsymbol{P}_o=(x_o,y_o)^T\in\mathbb{R}^{2\times1}$ is a pixel in the original pinhole image and $\boldsymbol{P}_d=(x_d,y_d)^T\in\mathbb{R}^{2\times1}$ is its corresponding pixel in distorted fisheye image. $[ k_1,k_2,k_3,k_4,...]$ are the radial distortion parameters and $r_f$ is the Euclidean distance between the distorted pixel and the distortion center $\boldsymbol{P}_c=(x_c,y_c)^T\in\mathbb{R}^{2\times1}$. 

Following the previous work of FisheyeEX~\cite{liao2022fisheyeex}, we use the same fourth order polynomial model to apply a distortion on original images. 
Besides, we acquire a circular mask by applying a mask generator. It masks out a circular region of the ground-truth image, aiming to make our synthesized fisheye image like the natural fisheye images captured via a fisheye camera.
Therefore, our datasets are composed of: complete-FoV fisheye images, complete-FoV fisheye semantic labels, and circular masks. The datasets distribution are shown in Fig.~\ref{fig:city_statis}.

\subsection{Implementation Details}
FishDreamer was implemented in PyTorch and trained for fisheye semantic completion end-to-end  on an NVIDIA RTX 3090 graphics card. Backbone weights are initialized from  models pretrained on ImageNet~\cite{deng2009imagenet}.
We choose the AdamW optimizer~\cite{loshchilov2017decoupled} with a learning rate of $2.5{\times}10^{-4}$, coefficients $\beta_{1}{=}0.9$, $\beta_{2}{=}0.999$, and weight decay $\eta{=}10^{-2}$. 
We use batch size of $8$ and train our model for $50$/$70$ epochs for the ablation experiments and the final model experiments respectively.
We adopt Peak Signal-to-Noise Ratio (PSNR), Structural Similarity Index Measure (SSIM), and Fréchet Inception Distance (FID) to  evaluate the extrapolation performance, while mean Intersection over Union (mIoU) is used as the semantic completion metric.

\begin{table}[!t]
\renewcommand{\thetable}{2}
    \begin{center}
        \caption{\emph{Quantitative comparison of fisheye semantic completion on Cityscapes-BF dataset.} * indicates results from~\cite{liao2022fisheyeex}.}
        \label{tab:cityscapes_result}
        \vspace{-1.0em}
        \resizebox{1.0\columnwidth}{!}{
\setlength{\tabcolsep}{1mm}{ 
\begin{tabular}{l|ccc:c}
\hline
\textbf{Dataset}   & \multicolumn{4}{c}{\textbf{Cityscapes-BF}}                                                \\

\hline

Sub-Task      & \multicolumn{3}{c:}{Image Completion}                       & Semantic Completion \\

\hline

Method    & PSNR $\uparrow$  & SSIM $\uparrow$ & FID $\downarrow$ & mIoU $\uparrow$   \\

\hline     
\hline  

SRN*~\cite{wang2019wide}       & 17.71 & 0.81 & 169.42 &      \textit{n.a.}                \\

RK*~\cite{liu2020rethinking}        & 21.79        & 0.87     &  136.66 &     \textit{n.a.}                     \\


HiFill*~\cite{yi2020contextual}     & 22.27        & 0.89     & 109.89   &      \textit{n.a.}                    \\


Boundless*~\cite{teterwak2019boundless} & 23.54        & 0.90     & 64.26   &      \textit{n.a.}                    \\


FisheyeEX*~\cite{liao2022fisheyeex} & 24.63        & \underline{0.92}     &      40.06   &      \textit{n.a.}               \\

\hline
FisheyeSeg~\cite{ye2020universal} & \textit{n.a.} &  \textit{n.a.} &  \textit{n.a.} & 47.31               \\
Swin-S + UPerNet & \textit{n.a.} &  \textit{n.a.} &  \textit{n.a.} & 53.65               \\
SegFormer-B2~\cite{xie2021segformer} &  \textit{n.a.} &  \textit{n.a.} &  \textit{n.a.} & 53.80               \\

\hline

Simple Baseline (Ours)      & \underline{24.82}      & \textbf{0.93}     &  \underline{34.48}      &     \underline{53.98}                \\

\rowcolor{gray!20}
FishDreamer (Ours)       & \textbf{25.05}      & \textbf{0.93} &  \textbf{30.14}      &     \textbf{54.54}                \\

\hline

\end{tabular}
}
}

        \vspace{-2em}
    \end{center}
\end{table}

\subsection{Results}
\begin{figure*}[!t]
   \centering
   \includegraphics[width=1.0\linewidth]{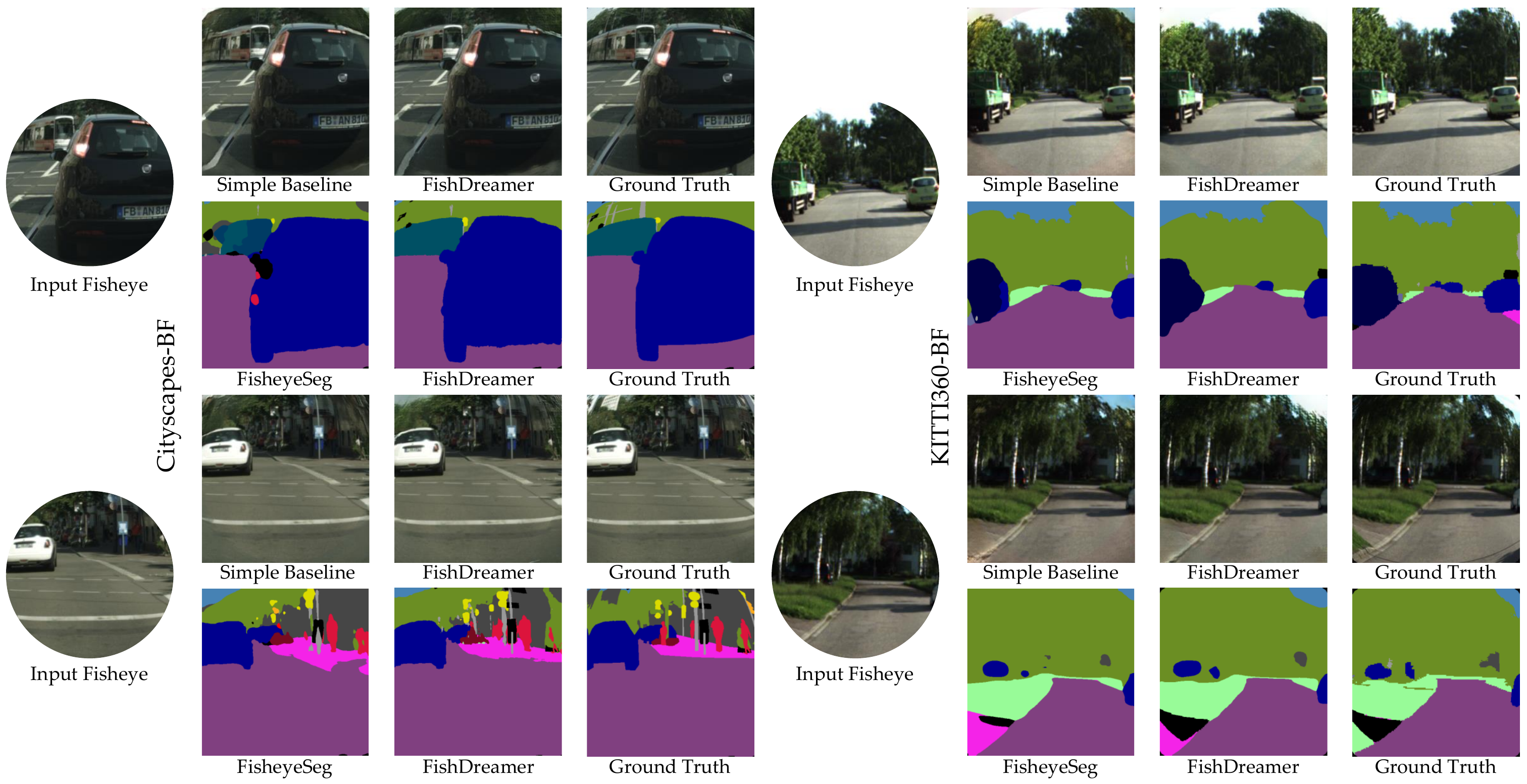}
   \vspace{-1.5em}
   \caption{\emph{Qualitative comparisons with alternative methods on Cityscapes-BF and KITTI360-BF. We compare the proposed FishDreamer with our simple baseline since it already surpasses previous works~\cite{liao2022fisheyeex} in terms of PSNR without the help of PCA. We also compare with the fisheye semantic segmentation method FisheyeSeg~\cite{ye2020universal}. Best viewed in color.}}
   \label{fig:compare}
  \vspace{-1.0em}
\end{figure*}

\begin{table}[!t]
\renewcommand{\thetable}{3}
    \begin{center}
        \caption{\emph{Quantitative comparison of fisheye semantic completion on KITTI360-BF dataset.} * indicates results from~\cite{liao2022fisheyeex}.}
        \label{tab:kitti_result}
        \vspace{-1.0em}
        \resizebox{1.0\columnwidth}{!}{
\setlength{\tabcolsep}{1mm}{ 
\begin{tabular}{l|ccc:c}
\hline
\textbf{Dataset}   & \multicolumn{4}{c}{\textbf{KITTI360-BF}}                                                \\

\hline

Sub-Task      & \multicolumn{3}{c:}{Image Completion}                       & Semantic Completion \\

\hline

Method    & PSNR $\uparrow$  & SSIM $\uparrow$ & FID $\downarrow$ & mIoU $\uparrow$   \\

\hline     
\hline  

SRN*~\cite{wang2019wide} & 18.25 & 0.79 & 143.90 &  \textit{n.a.}                \\

RK*~\cite{liu2020rethinking} & 20.13 & 0.82 & 102.77 &  \textit{n.a.}                     \\


HiFill*~\cite{yi2020contextual} & 20.10 & 0.83 & 82.61 &  \textit{n.a.}                     \\


Boundless*~\cite{teterwak2019boundless} & 21.52 & 0.87 & 53.17 &  \textit{n.a.}                    \\


FisheyeEX*~\cite{liao2022fisheyeex} & 22.31 & \underline{0.90} & 34.68 &  \textit{n.a.}              \\

\hline

FisheyeSeg~\cite{ye2020universal} & \textit{n.a.} &  \textit{n.a.} &  \textit{n.a.} & 39.08               \\

Swin-S + UPerNet &  \textit{n.a.} &  \textit{n.a.} &  \textit{n.a.} & 40.00               \\ %

SegFormer-B2~\cite{xie2021segformer} &  \textit{n.a.} &  \textit{n.a.} &  \textit{n.a.} & 41.19               \\

\hline

Simple Baseline (Ours) & \underline{22.38} & \underline{0.90} & \underline{30.23} & \underline{42.24}                \\

\rowcolor{gray!20}
FishDreamer (Ours) & \textbf{22.51} & \textbf{0.91} & \textbf{27.89} & \textbf{43.57}                \\

\hline

\end{tabular}
}
}

        \vspace{-3em}
    \end{center}
\end{table}

\noindent \textbf{Cityscapes-BF results.} In Table~\ref{tab:cityscapes_result}, we conduct comparison between the state-of-the-art methods for image completion, \eg, FisheyeEX~\cite{liao2022fisheyeex}, those for semantic segmentation, \eg, SegFormer~\cite{xie2021segformer}, and the proposed FishDreamer approach. Compared to the previous FisheyeEX, FishDreamer obtains better results on the sub-task of image completion, which are respective $25.05$, $0.93$, and $30.14$ in PSNR, SSIM, and FID. 
On the sub-task of semantic completion,  FishDreamer obtains the best score of $54.54\%$ in mIoU, yielding a large performance boost in ${+}7.23\%$ as compared to the previous fisheye image semantic segmentation model FisheyeSeg~\cite{ye2020universal}. Besides, compared to the methods for general image semantic segmentation, such as Swin~\cite{liu2021swin} and SegFormer~\cite{xie2021segformer}, our FishDreamer model also yields considerable improvement.
As the completion involves severe distortions and demands inferring semantics beyond the FoV, the segmentation transformers deliver clearly lower scores compared to their performances on standard segmentation benchmarks.
Yet, the state-of-the-art performance achieved in both tasks of FishDreamer demonstrates that outpainting the content outside the fisheye camera's FoV and completing the semantics of objects are conjugated. In other words, by using a single end-to-end model for both problems we can effectively leverage this complementary information, yielding clear benefits for both tasks.

\noindent \textbf{KITTI360-BF results.} 
As shown in Table~\ref{tab:kitti_result}, the results on the KITTI360-BF dataset are compared among fisheye image completion methods, semantic completion methods, and our two fisheye semantic completion approaches. Compared to image completion methods like Boundless~\cite{teterwak2019boundless} and FisheyeEX~\cite{liao2022fisheyeex}, our simple baseline achieves the second-best result ($22.38$) with significant improvement. Moreover, our FishDreamer model attains the best results in the image completion sub-task, with respective scores of $22.51$, $0.91$, and $27.89$ in PSNR, SSIM, and FID. In addition to image completion, our FishDreamer achieves the best semantic completion performance with an mIoU of $43.57$. These results and improvements further demonstrate the promising performance of our proposed method, which effectively couples both sub-tasks in an end-to-end manner.

\subsection{Ablation Studies}

\begin{table}[!t]
\renewcommand{\thetable}{4}
    \begin{center}
        \caption{\emph{Ablation of different backbones.}}
        \label{tab:ab_backbone}
        \vspace{-1.0em}
        \resizebox{1.0\columnwidth}{!}{
\setlength{\tabcolsep}{1mm}{ %
\begin{tabular}{lcccc}

\hline

\textbf{Backbone} & \textbf{PSNR} & \textbf{SSIM} & \textbf{mIoU} & \textbf{\#Params(M)}\\

\hline     
\hline  

Conformer-T~\cite{peng2021conformer}       & 24.34 & 0.9227 & 47.36 & \underline{29.0}              \\

Conformer-S~\cite{peng2021conformer}       & \textbf{24.57} & \textbf{0.9237} & 49.81 & 45.0              \\ 

\hline

MiT-B0~\cite{xie2021segformer}       & 23.28 & 0.9203 & 44.83 & \textbf{9.3}              \\

MiT-B2~\cite{xie2021segformer}       & 24.12 & \underline{0.9229} & 50.72 & 30.9              \\
\hline  
Swin-T~\cite{liu2021swin}       & 24.21 & 0.9167 & \underline{50.96} & 34.9               \\ 

\rowcolor{gray!20}
Swin-S~\cite{liu2021swin}       & \underline{24.46} & 0.9224 & \textbf{54.01} & 56.2               \\

\hline

\end{tabular}
}
}

        \vspace{-1.95em}
    \end{center}
\end{table}

\noindent \textbf{Analysis of the backbones.}
To investigate the effect of model backbones, we perform ablation study of FishDreamer with three different methods, including Swin~\cite{liu2021swin}, MiT~\cite{xie2021segformer}, and Conformer~\cite{peng2021conformer}. As shown in Table~\ref{tab:ab_backbone}, each method has two model scales. The best and the second best results are marked with bold and underline, respectively. Our method based on Conformer~\cite{peng2021conformer} models have better performance on the sub-task of fisheye image completion, and the model based on Conformer-S  achieves respective $24.57$ and $0.9237$ scores in PSNR and SSIM, but obtains sub-optimal performance in semantic completion, yielding only $49.81$ in mIoU. Compared to the Conformer models, based on MiT-B0 and -B2~\cite{xie2021segformer} backbones that are specific for semantic segmentation, our method achieves better results on the sub-task of semantic completion with $44.83$ and $50.72$ scores in mIoU. The MiT-based models have a smaller number of parameters ($9.3$M and $30.9$M), however, the performance on the fisheye image completion is  lower as compared to the ones using Conformer counterparts. 
To achieve a balance between the two sub-tasks, the Swin-based~\cite{liu2021swin} backbone  strikes a good balance between the fisheye image completion performance and the semantic completion quality.
Our method based on Swin-S backbone obtains the best semantic completion result with $54.01$ in mIoU, while it provides the second best result on fisheye image completion with $24.46$ in PSNR.
This result aligns with our observation that a backbone with superior semantic completion capabilities can provide complementary advantages for the image completion sub-task.
\begin{table}[!t]
\renewcommand{\thetable}{5}
    \begin{center}
        \caption{\emph{Analysis of Polar-aware Cross Attention.}}
        \label{tab:ab_pca}
        \vspace{-1.0em}
        \resizebox{1.0\columnwidth}{!}{
\setlength{\tabcolsep}{2mm}{ 
\begin{tabular}{lcccc}

\hline
\textbf{Polar Mask} & \textbf{Direction} & \textbf{mIoU} & \textbf{PSNR} & \textbf{SSIM}\\

\hline     
\hline  

\multirow{3}{*}{$w/o$ }      & S2P & 53.73 & \underline{24.93} & 0.9246               \\

      & P2S & \underline{54.17} & 24.56 & 0.9215 \\ 

       & Bi-direction & 53.89 & 24.82 & 0.9240 \\
\hline  

$2$       & Bi-direction & 54.10 & \underline{24.93} & \underline{0.9249}              \\

\rowcolor{gray!20}
$4$       & Bi-direction & \textbf{54.21} & \textbf{25.01}  & \textbf{0.9257} \\

$8$       & Bi-direction & 53.84 & 24.82 & 0.9242 \\
\hline  

\end{tabular}
}
}

        \vspace{-1.95em}
    \end{center}
\end{table}

\noindent \textbf{Analysis of the Polar-aware Cross Attention (PCA).}
The  PCA mechanism is vital for the fisheye semantic completion task.
To examine the impact of mask selection and direction, we perform an ablation study of PCA  in Table~\ref{tab:ab_pca}.
Without using the polar mask, the three ways of semantic-to-outpainting (S2P), outpainting-to-semantic (P2S), and Bi-direction achieve respective $53.73$, $54.17$, and $53.89$ in mIoU of semantic completion task, and $24.93$, $24.56$, and $24.82$ in image completion PSNR.
When using the polar mask and Bi-direction method, we further ablate the mask generation with different mask numbers in $\{2, 4, 8\}$. As shown in Table~\ref{tab:ab_pca}, our PCA module is robust to different mask numbers, since each of them obtains comparable performance.
Nonetheless, we found that using $4$ polar masks could provide better results on both sub tasks, yielding $54.21$, $25.01$, and $0.9257$ in mIoU, PSNR, and SSIM, respectively.
This analysis demonstrates that our proposed PCA module is effective in simultaneously addressing semantic understanding and image completion.

\section{Conclusion}
In this paper, we look into fisheye semantic completion, a novel task addressing the  extended field of view  perception in terms of simultaneous  image sensing and semantic understanding.
To tackle this challenge, we propose FishDreamer to intertwine image outpainting and segmentation via polar-aware cross attention, which guides outpainting with extended semantic contextual information in the annular dimension.
We establish Cityscapes-BF and KITTI360-BF benchmarks to assess the effectiveness of driving scene understanding beyond fisheye vision.
Extensive experiments demonstrate the effectiveness of the proposed polar-aware cross attention and FishDreamer over its counterparts.

In this future, we intend to introduce conditional diffusion models to enhance the completion and investigate the generalization of FishDreamer on real fisheye images.


\clearpage
{\small
\bibliographystyle{ieee_fullname}
\bibliography{egbib}
}

\end{document}